\documentclass[10pt,journal,compsoc]{IEEEtran}
\usepackage[english]{babel}
\usepackage{ragged2e}

\usepackage{epsfig}
\usepackage{graphicx}
\usepackage{subfigure}
\usepackage{algorithm}
\usepackage{algorithmic}

\ifCLASSOPTIONcompsoc
  \usepackage[nocompress]{cite}
\else
  \usepackage{cite}
\fi
\ifCLASSINFOpdf
\else
\fi
\begin{document}

\title{Compact Descriptors for Video Analysis: the Emerging MPEG Standard}
\author{Ling-Yu~Duan,  Vijay~Chandrasekhar, Shiqi~Wang, Yihang~Lou, Jie~Lin, Yan~Bai,
	Tiejun~Huang, Alex~Chichung~Kot,~\IEEEmembership{Fellow,~IEEE} and Wen~Gao,~\IEEEmembership{Fellow,~IEEE}
\thanks{Ling-Yu Duan and Vijay Chandrasekhar are
joint first authors.}
}

%
\maketitle
%
%

\begin{abstract}
This paper provides an overview of the on-going compact descriptors for video analysis standard (CDVA) from the ISO/IEC moving pictures
experts group (MPEG). MPEG-CDVA targets at defining a standardized bitstream syntax to enable interoperability
in the context of video analysis applications. During the developments of MPEG-CDVA, a series of techniques aiming to reduce the descriptor size and improve the video representation ability have been proposed. This article describes the new standard that is being developed and reports the performance of these key technical contributions.
\end{abstract}

\section{Introduction}
\label{sec:intro}

\IEEEPARstart{O}{ver} the past decade, there has been an exponential increase in the demand for video analysis, which refers to the capability of automatically analyzing the video content for event detection, visual search, tracking, classification, etc. Generally speaking, a variety of applications can benefit from the automatic video analysis, including mobile augmented reality (MAR), automotive, smart city, media entertainment, etc. For instance, MAR requires object recognition and tracking in real-time for accurate virtual object registration. With respect to automotive, robust object detection and recognition are highly desirable for warning the collision and cross-traffic. The increasing proliferation of surveillance systems is also driving the developments of object detection, classification and visual search technologies. Moreover, a series of new challenges have been brought forward in media entertainment, such as interactive advertising, video indexing and near duplicate detection, which all rely on robust and efficient video analysis algorithms.

For the deployment of video analysis functionalities in real application scenarios, a unique set of challenges are presented~\cite{mpeg_cdva_req}.
Basically, it is the central server which performs automatic video analysis tasks, such that efficient transmission of the visual data via a bandwidth constrained network is highly desired \cite{girod2011mobile}\cite{ji2011learning}. The straightforward way is to encode the video sequences and transmit the compressed visual data over the networks. As such, features can be extracted from the decoded videos for video analysis purpose. However, this may create high-volume data due to the pixel level representation of the video texture. One could imagine that 470,000 closed-circuit television (CCTV) cameras for video acquisition are deployed in Beijing, China. Assuming that for each video 2.5 Mbps bandwidth\footnote{2.5Mbps is the standard bitrate for 720P video with standard frame rate (30fps).} is required to ensure that they can be simultaneously uploaded to the server side for analysis, in total 1.2Tbps video data are transmitted on the internet highway for security and safety applications. Due to the massive CCTV camera deployment in the city, it is urgently required to investigate ways to handle the large scale video data.
\begin{figure}[]
\centering
\includegraphics[width=3.5in]{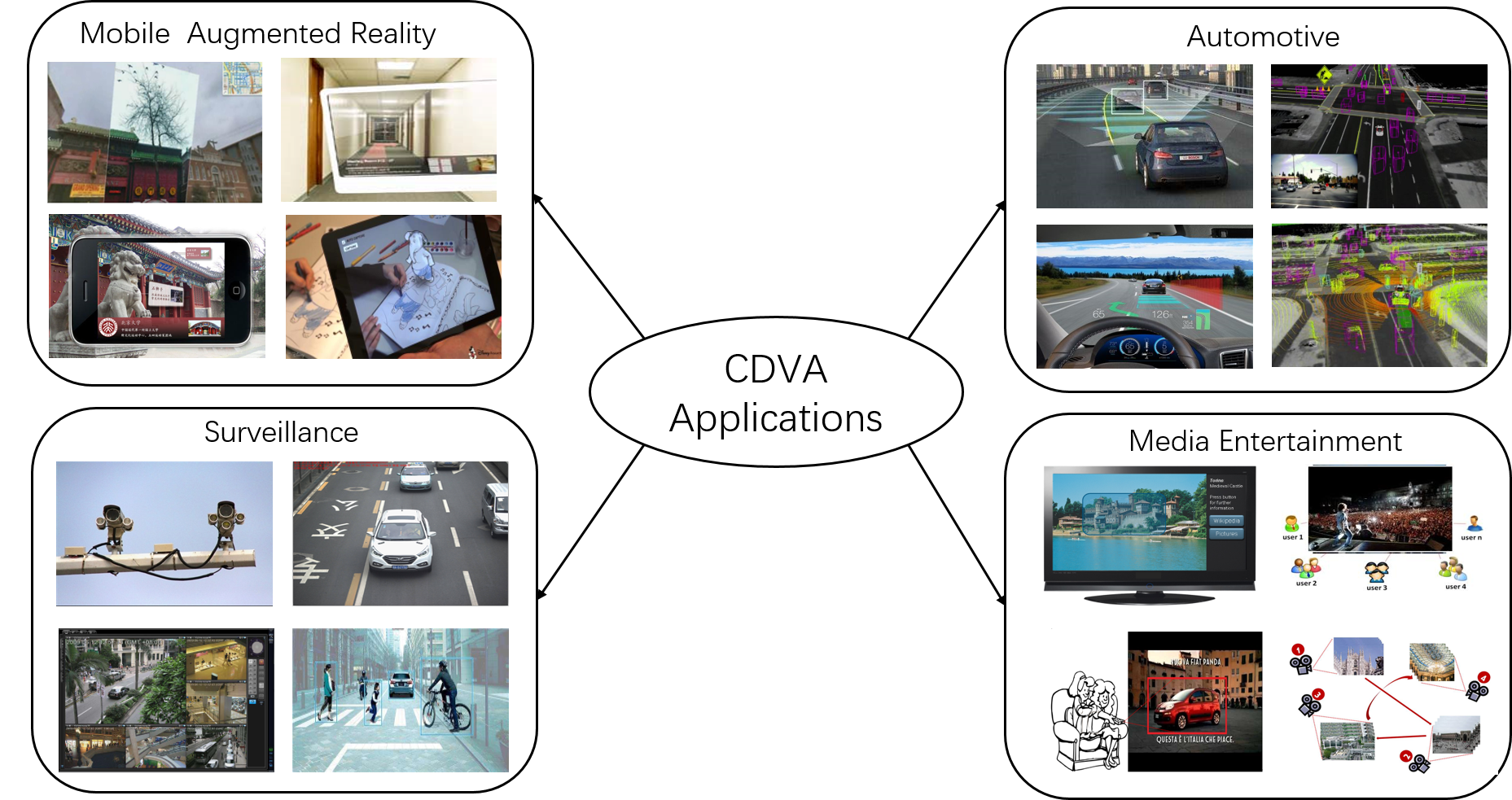}
\caption{Potential application scenarios of CDVA. }
\label{fig:app}
\end{figure}
%
%
%

As video analysis is directly performed based on extracted features instead of the texture, shifting the feature extraction and representation into the camera-integrated module is highly desirable, which directly supports the acquisition of features at the client side.
As such, compact feature descriptors instead of compressed texture data can be delivered, which can completely satisfy the requirements of video analysis. Therefore, developing effective and efficient compact feature descriptor representation techniques with low complexity and memory cost is the key to such ``analyze then compress'' infrastructure \cite{baroffio2013Compress}. Moreover, the inter-operability should also be maintained to ensure that feature descriptors extracted by any devices and transmitted in any network environments are fully operable at the server end.
The Compact Descriptors for Visual Search (CDVS) standard \cite{mpeg_cdvs_release}\cite{duan2016overview} developed by motion picture
experts group (MEPG), standardizes the descriptor bitstream syntax and the corresponding extraction operations of still images to ensure inter-operability for visual search applications. %
It has been proven to achieve high efficiency and low latency mobile visual search \cite{duan2016overview}, and an order of magnitude data reduction is realized by only sending the extracted feature descriptors to the remote server.

\begin{figure*}[]
\centering
\includegraphics[width=6.8in]{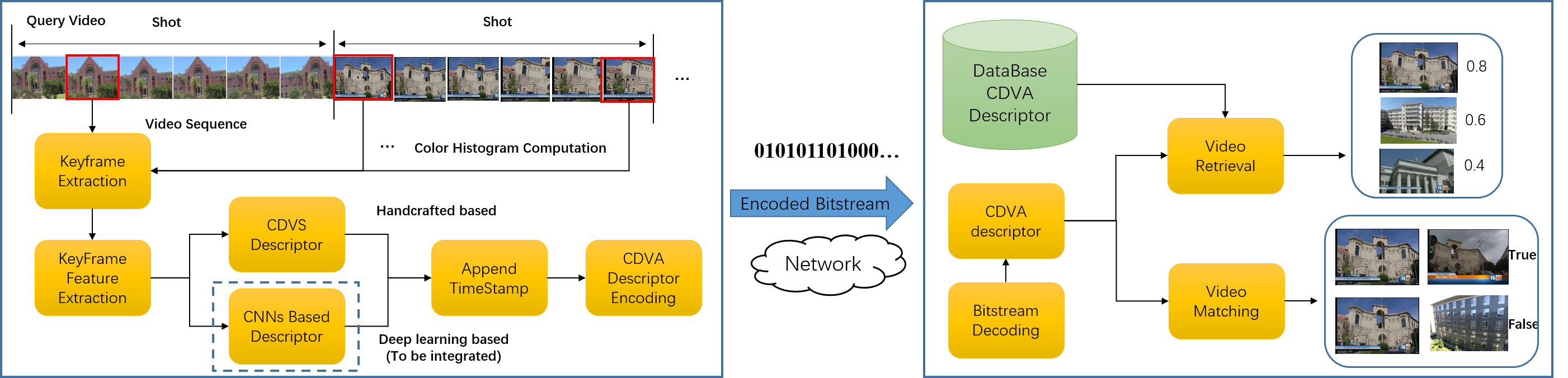}
\caption{Illustration of the CDVA framework.
}
\label{fig:cdva_pipeline}
\end{figure*}
However, the straightforward encoding of CDVS descriptors extracted frame
by frame from video sequences cannot fulfil the applications of video analysis.
For example, as suggested by CDVS, the descriptor length for each frame is 4K, and for a typical 30fps video the feature bit rate is
approximately to be 1Mbps. Obviously, this may lead to
excessive consumption of storage and bandwidth. Unlike still
images, video combines a sequence of high correlated frames
to form a moving scene. To fill the gap between the existing MPEG technologies and the emerging requirements of video feature descriptor compression, a Call for Proposals (CfP) on Compact Descriptors for Video Analysis (CDVA)~\cite{mpeg_cdva_cfp} was issued in 2015 by MPEG, targeting at enabling efficient and inter-operable design of advanced tools to meet the growing demand of video analysis. It is also envisioned that CDVA can achieve significant savings in memory size and bandwidth resources, and meanwhile provide hardware-friendly support for the deployment of CDVA at application level. As such, the aforementioned video analysis applications such as MAR, automotive, surveillance and media entertainment can be flexibly supported by CDVA~\cite{mpeg_cdva_use_case}, as illustrated in Fig.~\ref{fig:app}.

In Fig.~\ref{fig:cdva_pipeline}
, the framework of CDVA is demonstrated, which is comprised of keyframe/shot detection, video descriptors extraction, encoding, transmission, decoding and video analysis against a large scale database. During the development of CDVA, a series of techniques have been developed for these modules. The key technical contributions of CDVA are reviewed in this paper, including the video structure, advanced feature representation, and video retrieval and matching pipeline. Subsequently, the developments of the emerging CDVA standard are discussed, and the performance of the key techniques is demonstrated. Finally, we discuss the relationship between CDVS and CDVA and look into the future developments of CDVA.

\section{The MPEG CDVS Standard}
\label{sec:review}

MPEG-CDVS provides the standardized description of feature descriptors and the descriptor extraction
process for efficient and inter-operable still image search applications.
Basically, CDVS can serve as the frame level video feature description, which inspires the inheritance of CDVS features in the CDVA exploration.
This section discusses the compact descriptors specified in CDVS, which are capable of adapting the network bandwidth fluctuations for the support of scalability with the predefined descriptor lengths: 512 bytes, 1K, 2K, 4K, 8K and 16K. 
%
\subsection{Compact Local Feature Descriptor}

The extraction of local feature descriptors is required to be completed in a low complexity and memory cost way. Obviously this is much more desirable for videos. The CDVS standard adopts the Laplacian of Gaussian interest point detector. The low-degree polynomial (ALP) approach is employed to compute the local response after Laplacian of Gaussian filtering. Subsequently, a relevance measure is defined to select a subset of feature descriptors, which is statistically learned based on several characteristics of local features including scale, peak response of the LoG, distance to image centre, etc.
Handcrafted SIFT descriptor is adopted in CDVS as the local feature descriptors, and a compact SIFT compression scheme achieved by transform followed with ternary scalar quantization is developed to reduce the feature size. 
This scheme is of low-complexity and hardware favorable due to fast processing (transform, quantization and distance calculation).
%
In addition to the local descriptors, location coordinates of these descriptors are also compressed for transmission.
%
In CDVS, the location coordinates are represented as a histogram consisting of a binary histogram map and a histogram counts array. The histogram map and counts array are coded separately by a simple arithmetic coder and a sum context based arithmetic coder \cite{location_coding}.

\subsection{Local Feature Descriptor Aggregation}
CDVS adopts the scalable compressed Fisher Vector (SCFV) representation for mobile image retrieval. In particular, the selected SIFT descriptors are aggregated to the fisher vector (FV) by assigning each descriptor to multiple Gaussians in a soft assignment manner. To compress the high dimensional FVs, a subset of Gaussian components in the Gaussian Mixture Model (GMM) are selected based on the their rankings in
terms of the standard deviation of each sub-vector. The number of selected Gaussian functions is dependent on the available coding bits, such that descriptor scalability is achieved to adapt to the available bit budget. Finally, one-bit scalar quantizer is applied to support fast comparison with Hamming distance.

\section{Key Technologies in CDVA}

Driven by the success of MPEG-CDVS, which provides a fundamental groundwork for the development of CDVA, a series of technologies have been brought forward. In CDVA, the key contributions can be categorized into the video structure, video feature description and video analysis pipeline. The CDVA framework specifies how the video is structured and organized for feature extraction, where key frame detection and inter feature prediction methods are presented. Subsequently, the deep learning based feature representation is reviewed, and the design philosophy and compression methods of the deep learning models are discussed. Finally, the video analysis pipeline which serves as the server side processing module is introduced.

\subsection{Video Structure}

Video is composed of a series of highly correlated frames, such that extracting the feature descriptors for each individual frame may be redundant and lead to unnecessary computational consumptions. In view of this, a straightforward way is to perform key frame detection, following which only feature descriptors of the key frames are extracted. In \cite{mpeg_cdva_16_02_cxm}, the global descriptor SCFV in CDVS is employed to compare the distance between the current frame and the previous one. In particular, if the distance is lower than a given threshold, indicating that it is not necessary to preserve the current frame for feature extraction, the current frame is dropped. However, one drawback of this method is that for each frame the SCFV should be extracted, which brings additional computational complexity. In \cite{mpeg_cdva_16_05_keyframe}, the color histogram instead of the CDVS descriptors is employed for the frame level distance comparison. As such, the SCFV descriptors in non-key frames do not need to be extracted. Due to the advantage of this scheme, it has been adopted into the CDVA experimentation model (CXM) 0.2 \cite{mpeg_cdva_16_05_cxm}. In \cite{mpeg_cdva_16_10_keyframe,mpeg_cdva_17_1_keyframe}, Bailer proposed to modify the segment produced by the color histogram. In particular, for each segment, the medoid frame of each segment is selected, and all frames within this segment that have lower similarity in terms of SCVF than a given threshold are further chosen for feature extraction.

The key-frame based feature representation has effectively removed the video temporal redundancy, resulting in low
bitrate query descriptor transmission. However, this strategy has largely ignored the intermediate information between two key-frames. In \cite{mpeg_cdva_16_05_inter}, it is interesting to observe that densely sampled frames can bring better video matching and retrieval performance at the expense of increased descriptor size.
In order to achieve a good balance between the feature bitrate and video analysis performance, the inter prediction techniques for local and global descriptors of CDVS have been proposed \cite{mpeg_cdva_16_05_inter,mpeg_cdva_17_1_inter,mpeg_cdva_16_10_inter}. Specifically, in \cite{mpeg_cdva_16_05_inter}, the intermediate frames between two key frames are denoted as the predictive frame (P-frame). In P-frame, the local descriptor is predicted by the multiple reference frame prediction. For those local descriptors which cannot find corresponding references, they are directly written into the bit-stream. For global descriptors in P-frame, for the component selected from both current and previous frames, the binarized sub-vector is copied from the
corresponding one in the previous frame to save coding bits. In \cite{mpeg_cdva_17_1_inter,mpeg_cdva_16_10_inter}, it is further demonstrated that more than 50\% compression rate reduction can be achieved by applying lossy compression of local descriptors, without significant influence on the matching performance. Moreover, it is demonstrated that the global difference descriptors can be efficiently coded using adaptive binary arithmetic coding as well.

\subsection{Deep Learning Based Video Representation}

Recently, due to the remarkable success of deep learning, numerous approaches have been presented to employ the Convolutional Neural Networks (CNNs) to extract deep learning features for image retrieval~\cite{babenko2015aggregating,tolias2015particular}. In the development of CDVA, the Nested Invariance Pooling (NIP) has been proposed to obtain the discriminative deep invariant descriptors
, and significant video analysis performance improvement over traditional handcrafted features has been observed. In this subsection, we will review the development of deep learning features in CDVA from the perspectives of deep learning based feature extraction, network compression, feature binarization and the combination of deep learning based feature descriptors with handcrafted ones.

\subsubsection{Deep Learning Based Feature Extraction}

Robust video retrieval requires the features to be scale, rotation and translation invariant.
The CNN models incorporate the local translation invariance by a succession of convolution and pooling operations. In order to further encode the rotation and scale invariance into CNN, motivated by the invariance theory, the NIP was proposed to represent each frame with a global feature vector \cite{dcc2017,CDVA_NIP2_OCT2016,CDVA_NIP1_OCT2016}. In particular, the invariance theory provides a mathematically proven strategy to obtain invariant representations with the CNNs. This inspires the improvement on the geometric invariance of deep learning features based on the pooling operations of the intermediate feature maps in a nested way. Specifically, given an input frame, it can be rotated with $R$ times, and for each time the \textit{pool5} feature maps ($W \times H \times C$) is extracted. Here, \(W\) and \(H\) denote the width and height of the map and
\(C\) is the number of feature channels. Based on the feature map, the multi-scale uniform region of interest (ROI) sampling is performed, resulting in the 5-D feature reforestation with dimension ($R \times S \times W^{'} \times H^{'} \times C$). Here, $S$ is the number of sampled ROIs in multi-scale region sampling. Subsequently, NIP performs a nested pooling over translations ($W^{'} \times H^{'}$), scales ($S$) and finally rotations ($R$).
Therefore, a $C$-dimensional global CNN feature descriptor can be generated. The performance of NIP descriptors can be further boosted by the PCA whitening~\cite{babenko2015aggregating,tolias2015particular}. To evaluate the similarity between two NIP feature descriptors, the cosine similarity function is adopted.

\subsubsection{Network Compression}

CNN models such as AlexNet~\cite{krizhevsky2012imagenet} and VGG-16~\cite{simonyan2014very} contain millions of neurons, which cost hundreds of MBs for storage. This creates great difficulties in video analysis, especially when the CNN models are deployed at the client side for feature extraction in the ``analyze then compress'' framework. Therefore, efficient compression model of the neural network is urgently required for the development of CDVA. In \cite{CDVA_NIPcom1_Jan2017,CDVA_NIPcom2_Apr2017}, both scalar and vector quantization (VQ) techniques using the Lloyd-Max algorithm are applied to compress the NIP model. The quantized coefficients are further coded with Huffman coding. Moreover, the model is further pruned to reduce the model size by dropping the convolutional layers.
It is shown that the compressed models which have two orders of magnitude smaller than the uncompressed models lead to negligible loss in video analysis.

\subsubsection{Feature Descriptor Compression}

The deep learning based feature descriptor generated from NIP is usually in float-point, which is not efficient for the subsequent feature comparison process. As hamming distance can facilitate effective retrieval especially for large video collections, the NIP feature binarization has been proposed for compact feature representation \cite{CDVA_NIPbin_Apr2017}. In particular, the one-bit scalar quantizer is applied to simply binarize the NIP descriptor. As such, much less memory footprint and runtime cost can be achieved with marginally degraded performance loss.

\subsubsection{Combination of Deep Learning Based and Handcrafted Features}
\begin{figure*}[]
\centering
\includegraphics[width=6.2in]{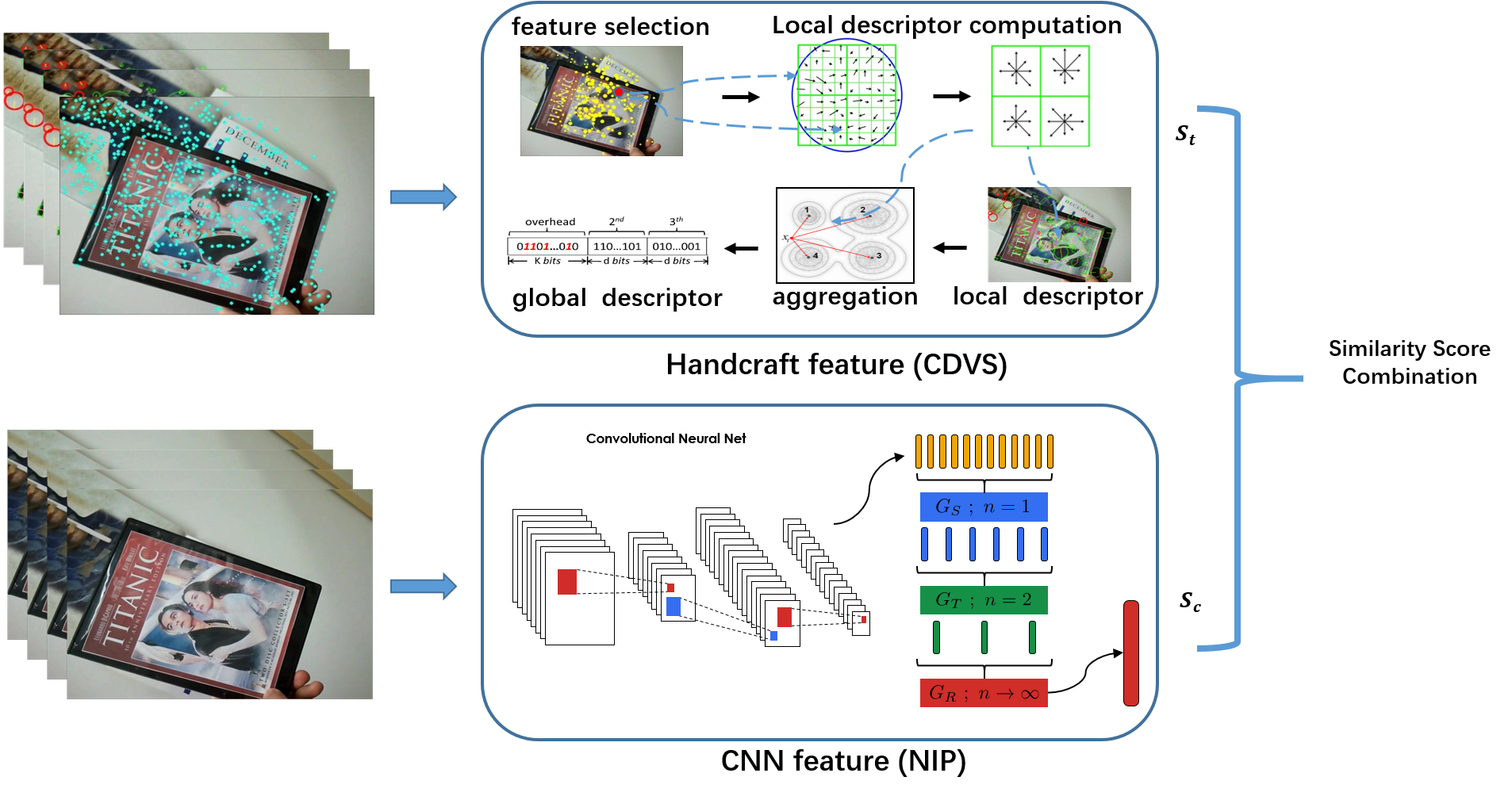}
\caption{Combination of handcrafted and deep learning based feature descriptors. }
\label{fig:combination}
\end{figure*}

Furthermore, in~\cite{CDVA_NIPcombination_Jan2017,dcc2017}, it is also revealed that there are some complementary effects between CDVS handcrafted and deep learning based features for video analysis. In particular, the deep learning based features are extracted by taking the whole frame into account while CDVS handcrafted descriptors sparsely sample the interest points. Moreover, the handcrafted features work relatively better in rich textured blobs, while deep learning based features are more efficient in aggregating deeper and richer features for global salient regions. Therefore, the combination of deep learning based features and CDVS handcrafted features has been further investigated in the CDVA framework \cite{CDVA_NIPcombination_Jan2017,dcc2017,CDVA_NIP1_OCT2016}, as shown in Fig.~\ref{fig:combination}. Interestingly, it is validated that the combination strategy achieves promising performance and outperforms either deep learning based or CDVS handcrafted features.

\subsection{Video Analysis Pipeline}

The compact description of videos enables two typical tasks in video analysis, including video matching and retrieval. In particular,
video matching aims at determining if a pair of videos shares the object or scene with similar content, and video retrieval performs searching for videos containing similar segment as the one in the query video.

\subsubsection{Video Matching}
Given the CDVA descriptors of the key frames in the video pair, pairwise matching can be achieved by comparing them in a coarse to fine strategy. Specifically, each keyframe in one video is first compared with all of the keyframes in the other video in terms of the global feature similarity. If the similarity is larger than the threshold, implying that there is a possible match between the two frames,
the local descriptor comparison can be further performed with the geometric consistency checking. The keyframe-level similarity is subsequently calculated by the multiplication of matching scores of the global and local descriptors. Finally, we can obtain the video-level similarity by selecting the largest matching score among all keyframe-level similarities.

Another criterion in video matching is the temporal localization, which locates the video segment containing similar items of interest based on the recorded timestamps. In \cite{CDVA_Localization_OCT2016}, a shot level localization scheme was adopted into CXM1.0. In particular, a shot is detected to be the group of consecutive
keyframes whose distance to the first keyframe of this shot is smaller than a certain threshold in terms of the color histogram comparison.
If the keyframe-level similarity is larger than a threshold, the shot that contains the key frame is regarded as the matching interval. Multiple matching intervals can also be concatenated together to obtain the final interval for localization.

\subsubsection{Video Retrieval}
In contrast to video matching, video retrieval is performed in a one-to-N manner, implying that the videos in the database are all visited and the top ones with higher matching scores are selected. In particular, the key-frame level matching with global descriptors is performed to extract the top $K_g$ candidate keyframes in the database. Subsequently, these key frames are further examined by local descriptor matching, and the key frame candidate dataset is further shrunk to $K_l$ according to the rankings in terms of the combination of global and local similarities. These key frames are reorganized into videos, which are finally ranked by the video level similarity following the principle in video matching pipeline.

\section{Emerging CDVA Standard}

\subsection{Evaluation Framework}

\begin{figure*}[]
\centering
\includegraphics[width=6in]{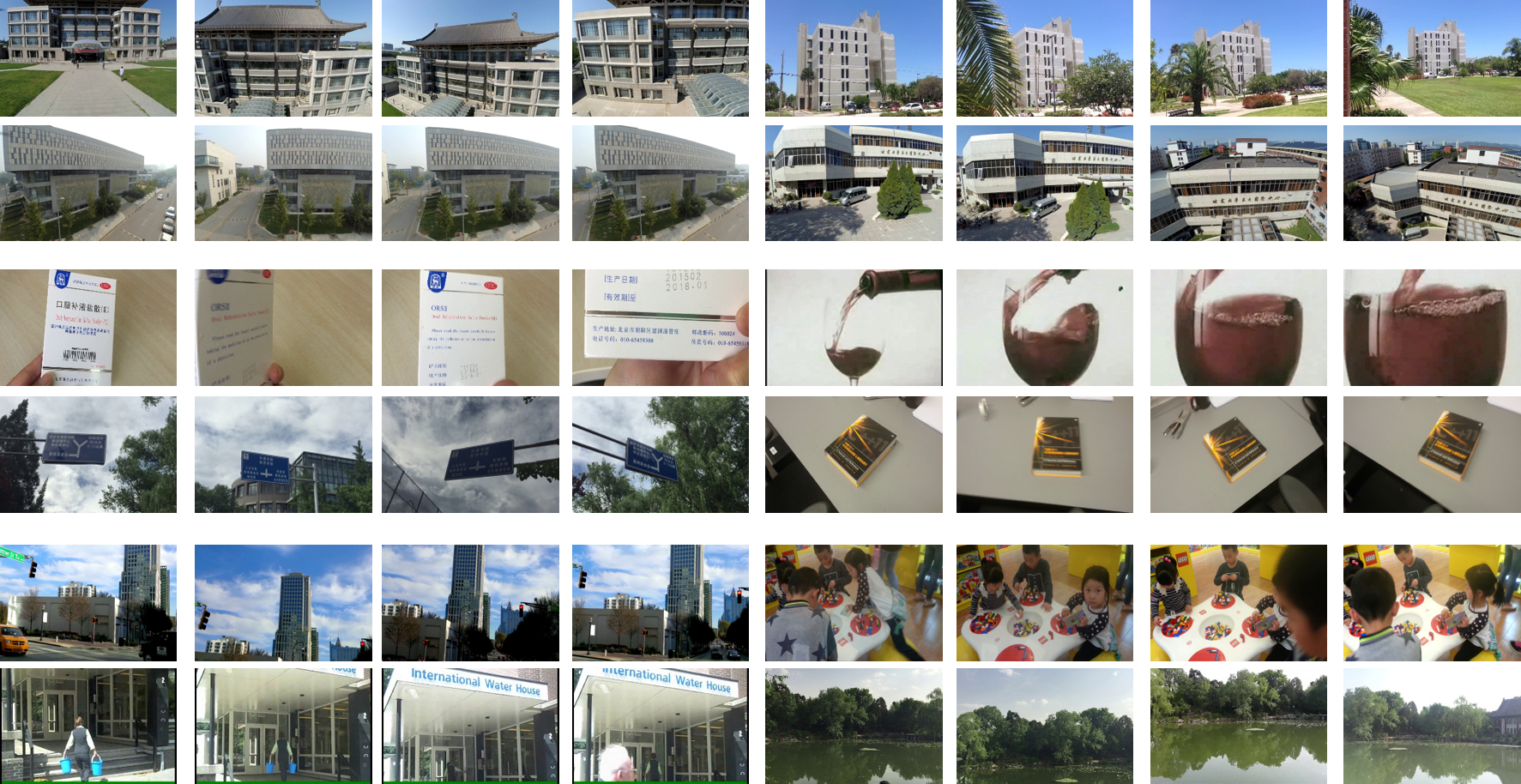}
\caption{Examples in the MPEG-CDVA dataset.}
\label{fig:CDVA_dataset}
\end{figure*}

The MPEG-CDVA dataset includes 9974 query and 5127 reference videos, and each video takes from 1 sec to 1+ min durations~\cite{mpeg_cdvs_eva}. In Fig.~\ref{fig:CDVA_dataset}, we provide some typical
examples from the MPEG-CDVA dataset.
In total, 796 items of interest in those videos are depicted, which can be further divided into three categories, including large objects (eg. buildings, landmarks), small objects (e.g. paintings, books, CD covers, products) and scenes (e.g. interior scenes, natural scenes, multi-camera shots). Approximately 80\% of query and reference videos were embedded in irrelevant content (different from those used in the queries). The start and end embedding boundaries were used for temporal localization in video matching task. The remaining 20\% of query videos were applied with 7 modifications (text/logo overlay, frame rate change, interlaced/progressive conversion, transcoding, color to monochrome and contrast change, add grain, display content capture) to evaluate the effectiveness and robustness of the compact video descriptor representation technique.
As such, 4,693 matching pairs and 46,930 non-matching pairs are created.
In addition, for large scale experiments 8,476 videos with a total duration of more than 1,000 hours are involved as distracters, which belong to UGC, broadcast archival and education.

The pairwise matching performance is evaluated in terms of the matching and localization accuracy. In particular, the matching accuracy is accessed by the
Receiver Operating Characteristic (ROC) curve. The True Positive Rate (TPR) given False Positive Rate (FPR) equaling to 1\% is also reported.
When a matching pair is observed, the localization accuracy is further evaluated by the Jaccard Index based on the temporal location of the item of interest within the video pair.
In particular, it is calculated by $\frac{[T_{start},T_{end}]\bigcap [T_{start}',T_{end}']}{[T_{start},T_{end}]\bigcup [T_{start}',T_{end}']}$,
where $[T_{start},T_{end}]$ denotes the ground truth and $[T_{start}',T_{end}']$ denotes the predicted start and end timestamps.
The retrieval performance is evaluated by mean Average Precision (mAP), and moreover precision at a given cut-off rank R for query videos (Precisian@R) is calculated. Here, R is set to be 100.
As the ultimate goal is to achieve compact feature representation, the feature bitrate consumption is also measured.

\subsection{Timeline and Core Experiments}
\begin{table*}[ht]
\centering
\caption{Timeline for the development of MPEG-CDVA standard.}
\label{my-label}
\begin{tabular}{|l|l|l|}
\hline
\textbf{When}  & \textbf{MileStone}                           & \textbf{Comments}                                                                                                                 \\ \hline
February, 2015 & Call for Proposals is published              & \begin{tabular}[c]{@{}l@{}}Call for Proposals for Compact Descriptor for\\ Video Analysis\end{tabular}                            \\ \hline
February, 2016 & Evaluation of proposals and Core experiments & None                                                                                                                              \\ \hline
October, 2017      & Draft International Standard                 & \begin{tabular}[c]{@{}l@{}}Complete specification. Only minor editorial changes\\ are allowed after DIS.\end{tabular}             \\ \hline
June, 2018     & Final Draft International Standard           & \begin{tabular}[c]{@{}l@{}}Finalized specification, submitted for approval\\ and publication.\end{tabular}                        \\ \hline
\end{tabular}
\end{table*}
The Call for Proposals of MPEG-CDVA was issued at the 111th
MPEG meeting in Geneva, in Feb. 2015, and
responses are evaluated in Feb. 2016. Table 1 lists the timeline for the development of CDVA.
In the current stage, there are six core experiments (CE) in the exploration of the MPEG-CDVA standard. The first CE investigates the temporal sampling strategy to better understand the impact of
key frames and densities in video analysis. The second CE targets at improving the matching and retrieval performance based on the segment level representation. The CE3 exploits the temporal redundancies of feature
descriptors to further reduce the bitrate for feature representation. CE4 investigates the combination strategy of traditional handcrafted and deep learning based feature descriptors, and CE5 develops compact representation methods of the deep learning based feature descriptors.
Finally, CE6 study the approaches for deep learning model compression to reduce the run time and memory footprint for deep learning based feature extraction.
\subsection{Performance Results}
In this subsection, we report the performance results of the key contributions in the development of CDVA. Firstly, the performance comparisons with the evolution of CXM models are presented. CXM 0.1 (released on MPEG-114) is the first version of CDVA experimentation model that provides the baseline performance, and subsequently CXM0.2 (MPEG-115) and CXM1.0 (MPEG-116) have been released. To flexibly adapt to different bandwidth requirements as well as application scenarios, three operating points in terms of the feature descriptor bit rate 16KBps, 64KBps and 256KBps are defined. Besides, in the matching operation, an additional cross mode 16\_256KBps matching has also been considered. In Table~\ref{tab:cxm}, the performance comparisons from CXM0.1 to CXM1.0 are listed. The performance improvements from CXM0.1 to CXM0.2 are significant, and more than 5\% on mAP and 5\% in terms of TPR@FPR are observed, which are mainly attributed to key frame sampling based on color histogram. Comparing CXM0.2 with CXM1.0, the retrieval performance is identical since the changes lie in the video matching operation, which improve the localization performance based on the video shot to identify the matching interval. Such matching scheme leads to more than 10\% temporal localization performance improvement.

\begin{table}[]
\centering
\caption{Performance comparisons with the evolution
of CXM models.}
\label{tab:cxm}
\begin{tabular}{|l|l|l|l|l|}
\hline
                                                                & \begin{tabular}[c]{@{}l@{}}Operating\\  Point\end{tabular} & CXM0.1 & CXM0.2 & CXM1.0 \\ \hline
mAP                                                             & 16KBps                                                        & 0.66   & 0.721  & 0.721  \\ \hline
                                                                & 64KBps                                                        & 0.673  & 0.727  & 0.727  \\ \hline
                                                                & 256KBps                                                       & 0.68   & 0.73   & 0.73   \\ \hline
Precisian@R                                                     & 16KBps                                                        & 0.655  & 0.712  & 0.712  \\ \hline
                                                                & 64KBps                                                        & 0.666  & 0.718  & 0.718  \\ \hline
                                                                & 256KBps                                                       & 0.674  & 0.722  & 0.722  \\ \hline
\begin{tabular}[c]{@{}l@{}}TPR@FPR\\ =0.01\end{tabular}         & 16KBps                                                        & 0.779  & 0.836  & 0.836  \\ \hline
                                                                & 64KBps                                                        & 0.79   & 0.843  & 0.843  \\ \hline
                                                                & 256KBps                                                       & 0.8    & 0.846  & 0.846  \\ \hline
                                                                & 16\_256KBps                                                   & 0.786  & 0.838  & 0.838  \\ \hline
\begin{tabular}[c]{@{}l@{}}Localization\\ Accuracy\end{tabular} & 16KBps                                                        & 0.365  & 0.544  & 0.662  \\ \hline
                                                                & 64KBps                                                        & 0.398  & 0.567  & 0.662  \\ \hline
                                                                & 256KBps                                                       & 0.411  & 0.579  & 0.652  \\ \hline
                                                                & 16\_256KBps                                                   & 0.382  & 0.542  & 0.652  \\ \hline
\end{tabular}
\end{table}

\begin{table*}[ht]
\centering
\caption{Performance comparisons between handcrafted and deep learning based methods.}
\label{tab:cxm_nip}
\begin{tabular}{|l|l|l|l|l|}
\hline
                             & mAP            & Precisian@R    & TPR@FPR=0.01   & \begin{tabular}[c]{@{}l@{}}Localization \\ Accuracy\end{tabular} \\ \hline
CXM0.1                       & 0.66           & 0.655          & 0.779          & 0.365                                                            \\ \hline
CXM0.2                       & 0.721          & 0.712          & 0.836          & 0.544                                                            \\ \hline
CXM1.0                       & 0.721          & 0.712          & 0.836          & 0.662                                                            \\ \hline
NIP                          & 0.768          & 0.736          & 0.879          & 0.725                                                            \\ \hline
\textbf{NIP+SCFV}            & \textbf{0.826} & \textbf{0.803} & \textbf{0.886} & \textbf{0.723}                                                   \\ \hline
NIP (compressed model)        & 0.763          & 0.773          & 0.87           & 0.722                                                            \\ \hline
NIP (compressed model) + SCFV & 0.822          & 0.798          & 0.878          & 0.722                                                            \\ \hline
Binarized NIP                & 0.71           & 0.673          & 0.86           & 0.713                                                            \\ \hline
Binarized NIP+ SCFV          & 0.799          & 0.775          & 0.872          & 0.681                                                            \\ \hline
\end{tabular}
\end{table*}

In Table~\ref{tab:cxm_nip}, the performance comparisons between CXM and deep learning based methods are provided.
Compared with CXM1.0, simply using the deep learning based feature descriptors in 512 dimension without re-ranking techniques can bring about 5\% improvements on both mAP and TPR.
It can be seen that the performance of NIP descriptor extracted from a compressed model only suffers a negligible loss while the model size has been reduced from 529.2M to 8.7M using pruning and scalar quantization. To meet the large-scale fast retrieval demand, the performance of binarized NIP (occupying only 512 bits) and its combination with handcrafted feature descriptors are also explored. Compared with CXM1.0, the additional 512 bits deep learning based descriptor in the combination mode significantly boosts the performance from 72.1\% to 79.9\%. It is worth noting that the results of deep learning method are under the cross-checking stage, and the Ad-hoc group plans to integrate the NIP descriptor into CXM in MPEG 119th meeting in Jul. 2017.

\begin{table}[]
\centering
\caption{Runtime complexity comprarisons between CXM1.0 and the deep learning based methods.}
\label{cxm:complexity}
\begin{tabular}{|l|l|l|l|}
\hline
                                                              & \begin{tabular}[c]{@{}l@{}}Retrieve\\ query(s/q.)\end{tabular} & \begin{tabular}[c]{@{}l@{}}Matching \\ pair(s/p.)\end{tabular} & \begin{tabular}[c]{@{}l@{}}Non Matching\\ piar(s/p.)\end{tabular} \\ \hline
CXM1.0                                                        & 38.63                                                          & 0.37                                                           & 0.21                                                              \\ \hline
NIP                                                           & 9.15                                                           & 0.3                                                            & 0.26                                                              \\ \hline
NIP+SCFV                                                      & 39.45                                                          & 0.38                                                           & 0.29                                                              \\ \hline
Binarized NIP                                                 & 2.89                                                           & 0.29                                                           & 0.25                                                              \\ \hline
\begin{tabular}[c]{@{}l@{}}Binarized NIP\\ +SCFV\end{tabular} & 33.24                                                          & 0.37                                                           & 0.28                                                              \\ \hline
\end{tabular}
\end{table}

In Table~\ref{cxm:complexity}, we list the runtime complexity between CXM1.0 and the deep learning based methods. In the experimental setup, for each kind of feature descriptor, the database is scanned once to generate the retrieval results. CXM1.0 adopts SCFV descriptor to obtain an initial top500 results and then local descriptor re-ranking is applied. The fastest method is binarized NIP that takes 2.89 seconds to implement a video retrieval request in 13603 videos (about 1.2 millon keyframes), and NIP descriptor takes 9.15 seconds to complete this task. For handcrafted descriptor, the CXM1.0 takes 38.63 seconds, including both global ranking with SCFV and re-ranking with local descriptors. It is worth mentioning that here CDVA mainly focuses on the performance improvement in terms of the accuracy of matching and retrieval. Regarding the retrieval efficiency, some techniques that have not been standardized in CDVS such as Multi-Block Index Table (MBIT)\cite{cdvs_mbit} indexing which can significantly improve the retrieval speed have not been integrated for investigation.

\section{Conclusions and Outlook}

The current development of CDVA treats the CDVS as the groundwork, as they serve the same purpose
of using compact feature descriptors for visual search and
analysis. The main difference lies in that CDVS is mainly
focusing on still images, while CDVA makes an extension to
video sequences. Moreover, the backward compatibility of CDVA supports
the feature decoding of the key frame with the existing
CDVS infrastructure, such that every standard compatible
CDVS decoder can reproduce the features of independently
coded frames in the CDVA bitstream. This can greatly facilitate the cross modality search applications, such as using images as queries to search videos, or using videos as queries to search corresponding images.

The remarkable technological progress in video feature representation has provided a further
boost for the standardization of compact video descriptors. The key frame representation and inter feature prediction
provide two granularity levels in video feature representation. The deep learning feature descriptors have also been intensively investigated, including the feature extraction, model compression, compact feature representation, and the combination of deep learned based features with
traditional handcrafted features. The optimization of the video matching and retrieval pipelines has also been proved to bring superior performance in video analysis.

Nevertheless, the standardization of CDVA is also facing many challenges and more improvements are expected. In addition to video matching and retrieval, more video analysis tasks (such as action recognition,
abnormal detection, video tracking) need to be investigated. This requires more advanced video representation techniques to extract the motion information as well as sophisticated deep learning models with high generalization abilty for feature extraction.
Moreover, although the deep learning method has achieved significant performance improvement, more deep feature compression and hashing work is necessary to achieve compact representation. Finally, the fusion strategy of deep learning feature
and traditional handcrafted feature pose new challenges to the standardization of CDVA and opens up new space for future
exploration.

\ifCLASSOPTIONcaptionsoff
  \newpage
\fi
\bibliographystyle{IEEEbib}
\bibliography{refs1,refs}

\end{document}